\documentclass[runningheads]{llncs}
\usepackage{booktabs,colortbl}
\usepackage{subcaption}
\usepackage[pdftex]{graphicx}
\usepackage{amsmath,bm}
\usepackage{amssymb}
\usepackage{cleveref}
\usepackage{wrapfig}
\usepackage{enumerate}
\usepackage[shortlabels]{enumitem}
\usepackage[misc]{ifsym}
\usepackage[hyphens]{url}

\begin{document}


\title{DiffusAL: Coupling Active Learning with Graph Diffusion for Label-Efficient Node Classification}

\titlerunning{DiffusAL: Diffusion-based Graph Active Learning}
%
\author{Sandra Gilhuber$^*$\inst{1,2}(\Letter) \and
Julian Busch$^*$\inst{1,3} \and
Daniel Rotthues\inst{1} \and\\
Christian M.M. Frey\inst{4} \and
Thomas Seidl\inst{1,2,4}}
\authorrunning{S. Gilhuber et al.}
%
\institute{LMU Munich, Germany  
\email{$\{$gilhuber,seidl$\}$@dbs.ifi.lmu.de} \and 
Munich Center for Machine Learning (MCML), Germany 
\and Siemens Technology, Princeton, NJ, USA \email{busch.julian@siemens.com}
\and Fraunhofer IIS, Germany 
\email{christianmaxmike@gmail.com}
}

\toctitle{DiffusAL: Coupling Active Learning with Graph Diffusion for Label-Efficient Node Classification}
\tocauthor{Sandra~Gilhuber}

\maketitle              
\def\thefootnote{*}\footnotetext{Equal contribution}\def\thefootnote{\arabic{footnote}}
\setcounter{footnote}{0}

\begin{abstract}
Node classification is one of the core tasks on attributed graphs, but successful graph learning solutions require sufficiently labeled data. To keep annotation costs low, active graph learning focuses on selecting the most qualitative subset of nodes that maximizes label efficiency. However, deciding which heuristic is best suited for an unlabeled graph to increase label efficiency is a persistent challenge.
Existing solutions either neglect aligning the learned model and the sampling method or focus only on limited selection aspects. They are thus sometimes worse or only equally good as random sampling. 
In this work, we introduce a novel active graph learning approach called \emph{DiffusAL}, showing significant robustness in diverse settings. Toward better transferability between different graph structures, we combine three independent scoring functions to identify the most informative node samples for labeling in a parameter-free way: i) \emph{Model Uncertainty}, ii) \emph{Diversity Component}, and iii) \emph{Node Importance} computed via graph diffusion heuristics. Most of our calculations for acquisition and training can be pre-processed, making DiffusAL more efficient compared to approaches combining diverse selection criteria and similarly fast as simpler heuristics. 
Our experiments on various benchmark datasets show that, unlike previous methods, our approach significantly outperforms random selection in 100\% of all datasets and labeling budgets tested. 
\keywords{active learning \and node classification \and graph neural networks}
\end{abstract}

\section{Introduction}
\label{sec:introduction}

Graph representation learning \cite{hamilton2020graph} and, especially, \textit{Graph Neural Networks} (GNNs) \cite{bronstein2017geometric,gilmer2017neural,battaglia2018relational} have been adopted as a primary approach for solving machine learning tasks on graph-structured data, including node classification~\cite{GCN}, graph classification~\cite{lee2018graph}, and link prediction~\cite{zhang2018link}. 
Applications range from quantum chemistry \cite{gilmer2017neural} over traffic forecasting \cite{zhao2019t} to cyber-security \cite{busch2021nf}.

However, supervised GNN models require sufficient training labels and usually assume that such labels are freely available. But, in reality, while unlabeled data is usually abundant, it is laborious and costly to provide annotations. 
Graph active learning has emerged as a promising direction to reduce labeling costs by carefully deciding which
data should be labeled to increase label efficiency. Under a limited budget, e.g., a fixed number of data samples to be labeled or time spent labeling by a domain expert, active learning aims to annotate an optimized set of training data iteratively. 
Hence, a key aspect of graph active learning is identifying the most informative instances in the abundance of unlabeled data for labeling. In particular, the goal is to be consistently more label-efficient than random labeling. Since random sampling is arguably the fastest and least complex method, active learning methods that are not significantly better than random sampling are not worthwhile.

However, since graphs can vary widely, it is very difficult to design an approach significantly better than random sampling across different labeling budgets and graph structures.  
Existing graph-active learning approaches reach their limits for various reasons:
Some approaches focus only on limited selection aspects~\cite{LSCALE,PreGEEM} and outperform random selection only on certain graphs. 
Others focus on one-shot selection without iterative re-training and active selection and can therefore not exploit model-related uncertainty scores~\cite{GRAIN,FeatProp}. 
Other methods try to include various criteria in the selection but are sensitive to user-defined hyper-parameters or are not deliberately aligned with the used model architecture~\cite{AGE,ANRMAB}. 
Moreover, many methods use a GCN\cite{GCN} for training and acquisition. However, GCNs learn latent node features and perform neighborhood aggregation in a coupled fashion, which can negatively influence the time needed for the active learning procedure. 
In contrast, \textit{Graph diffusion} is a promising direction tackling limitations such as restriction to \textit{k}-hop neighborhoods \cite{PushNet} or over-smoothing, where neighborhood aggregation and learning are decoupled.

\begin{figure}[t]
    \centering
    \includegraphics[width=\textwidth,keepaspectratio]{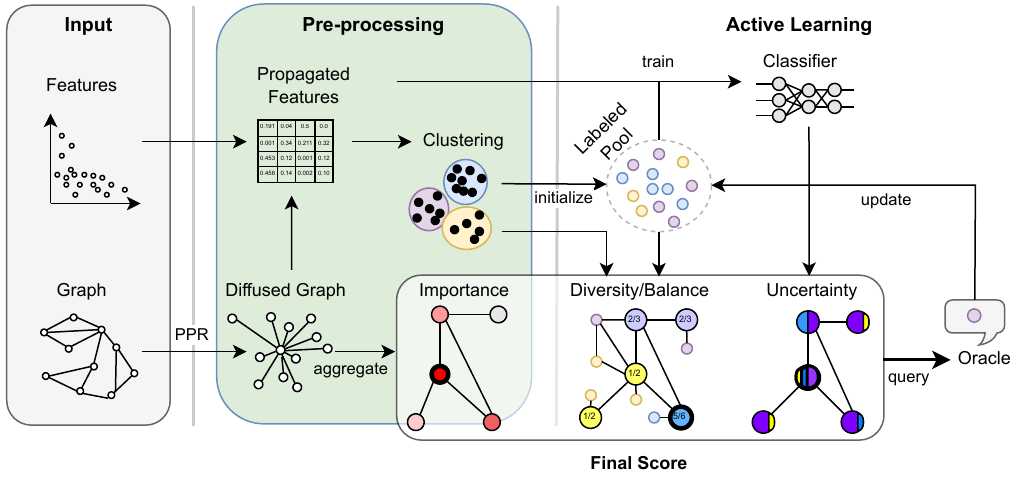}
    \caption[DiffusAL general pipeline]{DiffusAL pipeline consisting of the original input graph and corresponding node features (grey box), pre-computed static model-independent scores, such as the propagated feature matrix and derived node importance (green box), a dynamic, model-independent score based on the composition of the labeled pool (Diversity/Balance), as well as a dynamic, model-dependent informativeness score (Uncertainty). These scores are combined into a final node rating (white box) to select the most useful instances for annotation.}
    \label{fig:contrib_pipeline}
\end{figure}

In this work, we use diffusion-based heuristics to combine graph learning with active learning. In particular, we propose \textit{DiffusAL}, a novel graph active learning method that leverages graph diffusion for highly accurate node classification and efficiently re-uses the computed diffusion matrix and diffused node feature vectors in the learning procedure. 

We introduce a new scoring function for identifying a node's utility which consists of three factors: i) \emph{Model Uncertainty}, ii) \emph{Diversity Component}, and iii) \emph{Node Importance}. 
DiffusAL combines these scores in a parameter-free scoring function that naturally adapts to consecutively learning iterations. 

Specifically, for i) \emph{Model Uncertainty}, we exploit a state-of-the-art scoring that has shown an improving impact on the selection of nodes \cite{SettlesAL}. Next, the ii) \emph{Diversity Component} refers to the variability of node features and, therefore, their respective labels. For that, we apply a clustering method on the pre-computed diffusion matrix where diversity is reached by picking samples from underrepresented communities. 
Finally, for computing iii) \emph{Node Importance}, we exploit the information given by diffusion matrix based on the Personalized PageRank (PPR), which provides information about the relative importance of nodes in a graph w.r.t. a particular seed node. 
The high-level key concepts of DiffusAL are illustrated in \Cref{fig:contrib_pipeline}.

We evaluate DiffusAL on five real-world benchmark datasets, demonstrating its superiority over a variety of competitors. Notably, DiffusAL is the only competitor to outperform random selection with statistical significance in 100\% of the evaluated datasets and labeling budgets. In a series of ablation studies, we show that DiffusAL works consistently well on all benchmark datasets, analyze which components contribute to its performance, and investigate its efficiency. 

In summary, our contributions are as follows:
\begin{itemize}
    \item 
    Enhancing the selection of influential nodes by using \emph{diffusion-based node importance} and utilizing pre-computed \emph{clustering on diffused features} to prevent oversampling a particular region.
    \item Combining three complementary node scoring components in a parameter-free way.
    \item Achieving high efficiency by propagating statically pre-computed features stored in a diffusion matrix.
\end{itemize}

\section{Related Work}
\label{sec:related_work}
Early works on graph active learning~\cite{bilgic2010active,moore2011active} exploit the graph structure for selecting nodes for querying without graph representation learning. 
More recent approaches~\cite{AGE,ANRMAB,LSCALE,PreGEEM,Wu_2019_nodefeatureKMeans} use GCNs to exploit the graph structure as well as learned features.
FeatProp~\cite{Wu_2019_nodefeatureKMeans} leverages node feature propagation followed by K-Medoids clustering for the selection of instances. By defining the pairwise node distances between the corresponding propagated node features, the model selects nodes being closest to the cluster representatives yielding a diverse set over the input space. However, the diversity scoring function in our model puts more weight on underrepresented clusters yielding a more balanced view of the available data space and, therefore, is more suitable for imbalanced data. In \cite{GRAIN}, the authors proposed \textit{GRAIN}, a model inspecting social influence maximization for data selection. Their objective is a diversified influence maximization by exploiting novel influence and diversity functions. 
In contrast to their work, we focus on an iterative active learning setting \cite{Contardo_2017_AMA} since it directly enables exploiting the uncertainty scores entangled to a model which is known to be valuable for query selection.
The most related work to our approach is presented in \cite{AGE} where the authors propose \textit{Active Graph Embedding} (AGE) using as selection heuristic a weighted sum of information entropy, information density, and graph centrality defined on direct neighborhoods. For the latter, they propose to use PageRank centrality. The time-sensitive coefficients of the weighted sum are chosen from a beta distribution using the number of training iterations as input. 
We overcome these limitations related the restriction on direct neighborhoods aggregations used in standard GNNs \cite{bronstein2017geometric,gilmer2017neural,battaglia2018relational} by leveraging continuous relationships via graph diffusion \cite{GDC,PushNet}.
In \cite{ANRMAB}, \emph{ANRMAB} is proposed. It uses a multi-armed bandit mechanism for adaptive decision-making by assigning different weights to different criteria when constructing the score to select the most informative nodes for labeling.  
The model \textit{LSCALE}~\cite{LSCALE} exploits clustering-based (K-Medoids) active learning on a designed latent space leveraging two properties: low label requirements and informative distances. For the latter, the authors integrate \textit{Deep Graph Infomax}~\cite{DGI} as an unsupervised model. Therefore, in contrast to our approach, the model utilizes a purely distance-based selection heuristic. 
The method \emph{GEEM} \cite{Regol_2020_geem} maximizes the expected error reduction to select informative nodes to label. 

To the best of our knowledge, we are the first to leverage the power of diffusion-based heuristics for the computation of node importance, being an integral part of our scoring function, combining three complementary components to compute the nodes yielding the highest utility scores. 
Moreover, our novel scoring function uncouples from any parameter presets, being a critical choice without any a priori knowledge about the input data. 

\section{DiffusAL}
\subsection{Preliminaries}
\paragraph{Notation.~}
\label{subsec:preliminaries_notation}
We consider the problem of active learning for node classification. We are given a graph $G = (V,E)$ represented by an adjacency matrix $A \in \{0, 1\}^{n \times n}$ along with a node feature matrix $X \in \mathbb{R}^{n \times d}$. Each node $v \in V$ belongs to exactly one class $c_v \in \{1, \dots, C\}$, where $C$ is the number of classes present in the dataset. A budget constraint $B$ denotes the maximum number of nodes for which the active learning algorithm may request the correct labels from the oracle. The main goal is to select a subset of nodes $S \subset V$ such that $|S| = B$ and the accuracy of a classification model trained on these nodes is maximized. In a batch setting, $b$ denotes the number of nodes selected within each acquisition round.

\paragraph{Recap: Feature Diffusion.~}
\label{subsec:preliminaries_featureDiff}

In contrast to conventional GNN architectures \cite{GCN,GAT,GIN} that learn latent node features and perform neighborhood aggregation in a coupled fashion, \textit{graph diffusion} effectively decouples the two steps to address certain shortcomings of conventional GNN architectures, including the restriction to \textit{k}-hop neighborhoods \cite{PushNet} and issues related to over-smoothing\cite{li2018deeper,xu2018representation,ogawa2021adaptive,Frey2022SEA}.
The general effectiveness of diffusion, when paired with conventional GNN architectures, was shown in \cite{GDC}. 
In general, a parametric diffusion matrix can be defined as 
\begin{equation}
    P = \sum_{k=0}^{\infty} \theta_k T^k,
\end{equation}
where $T$ is a transition matrix and $\theta$ are weighting parameters. A popular choice is \textit{Personalized PageRank (PPR)} \cite{faerman2018semi,borutta2019structural,APPNP,faerman2020ada,PushNet}, where $T=AD^{-1}$ is the random walk matrix, $D$ is the diagonal degree matrix, and $\theta_k = \alpha(1-\alpha)^k$. Intuitively, $P_{ij}$ corresponds to the probability that a random walk starting at node $i$ will stop at node $j$ and can be interpreted as the importance of node $j$ for node $i$. The restart probability $\alpha \in [0, 1]$ controls the effective size of a node's PPR-neighborhood.
An approximation of the PPR matrix can be pre-computed in time $O(n)$ using push-based algorithms \cite{PushNet}. This approximation requires a second hyper-parameter $\varepsilon > 0$ that thresholds small entries and, thus, has a sparsification and noise reduction effect.
Once computed, the PPR matrix can replace the adjacency matrix used by conventional message-passing networks for neighborhood aggregation \cite{APPNP,PushNet}. 

\subsection{Model architecture}%
\label{subsec:diffusal_architecture}%
For DiffusAL, we propagate the original node features such that the propagated node features don't depend on any learned transformations and can be pre-computed as well. We propose a query-by-committee (QBC) approach \cite{QBC}, where the propagated node features are connected to an ensemble of MLP classifiers to robustify uncertainty estimation during the sample selection process compared to a commonly used single MLP.
Additionally, features are diffused over multiple scales by varying the hyper-parameter $\alpha$ controlling the effective neighborhood size over which features are aggregated. In particular, the model predictions are given as 
\begin{equation}
    Y = predict \left( \sum_{j \in \{1, \dots, M\}} transform_j \left( \sum_{i \in \{1, \dots, K\}} P^{(\alpha_i)} X \right) \right),
\end{equation}
where $K$ denotes the number of scales, and $M$ denotes the number of MLPs in the classification ensemble. The pre-computed diffused features are aggregated over multiple scales using the $sum$ function and fed to the hidden layer of each MLP. The learned representations are then aggregated using the $sum$ function and passed to the shared prediction layer. All ensemble members share the same architecture and only differ in the random initialization of their weights and biases. The QBC can be trained very efficiently with gradient descent, and, in particular, the expensive diffusion step needs to be performed only once as a pre-processing step.

\subsection{Node Ranking and Selection}%
\label{subsec:node_scoring}%
In addition to facilitating highly effective and efficient prediction, the previously computed diffusion matrix $P = \sum_{i \in \{1, \dots, K\}}P^{(\alpha_i)}$ and diffused features $PX$ are reused to calculate expressive ranking scores for active node selection.

\leavevmode\\
\textbf{Model Uncertainty.~}
For measuring model uncertainty, we utilize the QBC defined above. In particular, we compute the Shannon entropy over the softmax-ed output distribution to determine the \textbf{uncertainty score} for node $i$:
\begin{equation}
    s_{\text{unc}}(i) = - \sum_{j \in \{1, \dots, C\}} y_{ij} \log y_{ij}.
\end{equation}
The scores are L1-normalized over all unlabeled nodes to $[0,1]$, so all scoring functions share the same scale and can be sensibly combined.

While this score is inspired by the classical query-by-committee~\cite{QBC} approach, it differs in the sense that it doesn't average the softmax outputs of the individual committee members but rather considers the softmax output of a single shared prediction layer applied to aggregated latent representations. Thereby, differing predictions become more distinct in the softmax output.

\leavevmode\\
\textbf{Diversity Component.~}%
For the diversity component, we perform $k$-Means clustering on the \emph{diffused features} with $k=b$ and assign each node a pseudo-label based on the clustering result. 
Note that we pre-compute these cluster assignments such that no re-computations are necessary at query time, in contrast to other approaches (e.g., based on GCNs), where updated node features would change the clustering.

The cluster-based pseudo labels are used to ensure decent coverage of the feature space. At each iteration, each node $i$ receives a \textbf{diversity score}
\begin{equation}
    s_{\text{div}}(i) = 1 - \frac{|c_{train}|}{|V_{train}|},
\end{equation}
where $c \in C$ denotes the cluster node $i$ was assigned to, $|c_{train}|$ denotes the number of nodes in the currently labeled training set belonging to cluster $c$, and $|V_{train}|$ is the number of currently labeled training nodes. In short, each node in the unlabeled pool is weighted by the relative size of its cluster in the training set, such that nodes from currently underrepresented clusters receive a higher score.
In contrast to only focusing on avoiding redundancy in the current batch~\cite{BADGE}, our diversity score can also be interpreted as a balancing score ensuring that no region is over-sampled within the labeled pool. 

Some existing works on graph active learning~\cite{AGE,ANRMAB} ignore the limitations of a randomly initialized labeled pool and ensure class balance. However, this simplification is rather unrealistic in a real-world active learning setting.
To overcome this limitation, we again exploit the $k$-Means clustering used for the diversity score and select nodes closest to centroids for the initial pool, inspired by clustering-based sampling approaches~\cite{FeatProp,LSCALE} and existing work on initial pool selection~\cite{InitialPools}.

\leavevmode\\
\textbf{Node Importance.~}
Graph diffusion allows for a natural way to quantify node importance. Since the weights $P_{ij}$ used for neighborhood aggregation can be interpreted as importance scores, summing up the importance of a node $i$ for all other nodes $j$ yields a measure of the general importance of node $i$, measuring its total influence on the predictions for other nodes. Since the columns of $S$ are stochastic, this procedure yields consistently scaled overall importance scores. In particular, the \textbf{importance score} of node $i$ is given by the row-wise sum
\begin{equation}
    s_{\text{imp}}(i) = \sum_{j \in V}P_{ij}.
\end{equation}
Since the importance scores for all nodes can be computed directly from the PPR matrix, they can be pre-computed before the active learning cycle starts.
Our node importance score is a proxy for how much influence a node has on other nodes, where nodes with higher scores are assumed to carry more valuable information about many other nodes as well. Node importance could be interpreted as a novel representativeness measure, which has been quantified via density- or center-based selection within previous (graph) active learning approaches \cite{AGE,ANRMAB,ALG}. However, we do not need to recompute a clustering on learned representations after each selected sample, nor do we require good representations since we can extract the information directly from the graph topology. Further, our importance score of a node directly reflects the influence of that node on the model's predictions, since the weights from which we compute the scores are directly used for neighborhood aggregation. This is not the case for alternative existing measures.

\begin{figure}[t]
    \begin{center}
        \begin{subfigure}[t]{0.46\textwidth}
            \centering
            \includegraphics[width=0.55\textwidth]{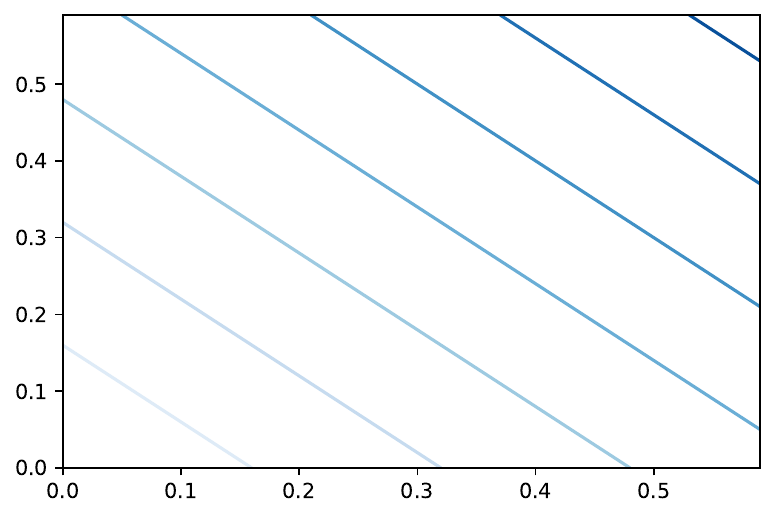}
            \caption{\textbf{Sum aggregation}: Isolines are straight due to fixed weighting.}
        \end{subfigure}
        \hfill
        \begin{subfigure}[t]{0.46\textwidth}
            \centering
            \includegraphics[width=0.55\textwidth]{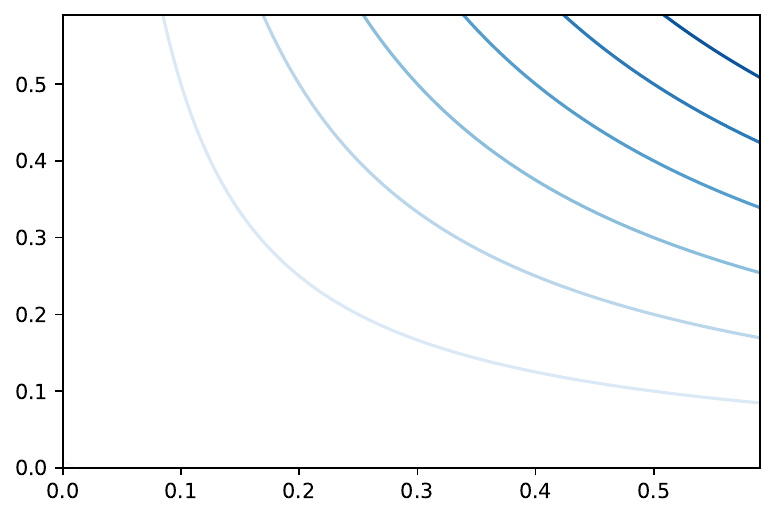}
            \caption{\textbf{Multiplicative aggregation}: Isolines are curved, favoring similar values over diverging ones.}
        \end{subfigure}

    \end{center}
    \caption{Score aggregation: for two arbitrary scores on the x and y axes (e.g. uncertainty and representativeness), the corresponding aggregated score is depicted as an isoline, i.e., each point on the line corresponds to the same final value.}
    \label{fig:score_aggregation}
\end{figure}

\leavevmode\\
\textbf{Score Combination and Node Selection.~}%
In summary, the uncertainty score assigns higher weights to nodes about which the committee is most uncertain, the diversity score assigns higher weights to nodes belonging to underrepresented clusters, and the node importance score assigns higher weights to nodes with a higher influence on the predictions for other nodes. The individual scores for a node are combined in a multiplicative fashion to determine the node's utility:
\begin{equation}
    s(i) = s_{\text{unc}}(i) \cdot s_{\text{div}}(i) \cdot s_{\text{imp}}(i).
\end{equation}
As illustrated in Figure \ref{fig:score_aggregation}, the intuition behind the multiplicative combination is to slightly favor nodes displaying a well-rounded distribution of scores over those with a strong imbalance when the sum of the scores is identical while still allowing extraordinarily important or uncertain nodes to be selected. 
Existing works use slightly different variations of time-sensitive weighted sums, thereby gradually shifting the focus from representativeness to uncertainty~\cite{AGE,ALG}. A disadvantage of time-sensitive weighting is that the performance of the selection algorithm depends on the choice of good hyper-parameters, which is difficult in a real-world active learning setting. In contrast, our multiplicative approach is parameter-free and naturally time-sensitive. In the early stages of training, the classifiers essentially guess predictions more or less uniformly, leading to roughly similar uncertainty scores for most nodes. Consequently, the uncertainty score is close to a constant factor applied equally to all nodes, thus naturally making the model-free scores the deciding ones in the final score. However, uncertainty scores become increasingly important once the classifiers become more confident in their predictions.
The combined utility score is determined for each unlabeled node in each active learning cycle. Afterward, the unlabeled nodes are ranked according to their utility, and the nodes with the highest utility scores are labeled.

\section{Experiments}
\label{sec:experiments}
To demonstrate the effectiveness and efficiency of DiffusAL, we conduct a series of experiments.
In particular, we investigate three research questions:
\begin{enumerate}[start=1,label={}]
    \item \textbf{R1} - How does DiffusAL perform compared to state-of-the-art methods?
    \item \textbf{R2} - How does each of DiffusAL's components contribute?
    \item \textbf{R3} - How is the training and acquisition efficiency?
\end{enumerate}

\begin{table}[t]
    \centering
    \caption[Dataset statistics]{Dataset statistics (only considering the largest connected component).}
    \setlength{\tabcolsep}{5pt}
    \begin{tabular}{lrrrr}
         \toprule
          \textbf{Dataset} & \textbf{{\#Nodes}} & \textbf{{\#Edges}} & \textbf{{\#Features}} & \textbf{\#Classes} \\ 
         \midrule
         \textbf{Citeseer} & 2120 & 3679 & 3703 & 6\\
         \textbf{Cora} & 2485 & 5069 & 1433 & 5\\
         \textbf{Pubmed} & 19717 & 44324 & 500 & 3\\
         \textbf{Co-author CS} & 18333 & 81894 & 6805 & 15\\
         \textbf{Co-author Physics} & 34493 & 247962 & 8415 & 5\\
        \arrayrulecolor[rgb]{0,0,0}
        \bottomrule
    \end{tabular}
    \label{table:dataset_stats}
\end{table}

\subsection{Experimental Setup}
\noindent\textbf{Datasets.~}
We evaluate DiffusAL on several well-established benchmark datasets for node classification, namely the citation networks Citeseer \cite{CiteSeerCora}, Cora \cite{CiteSeerCora} and Pubmed \cite{PubMed}, as well as the co-author networks Computer Science (CS)~\cite{Coauthor} and Physics~\cite{Coauthor}, summarized in \Cref{table:dataset_stats}. For each dataset, only the largest connected component is used, and features are L1-normalized.

\noindent\textbf{Implementation Details.~} 
All experiments were implemented using PyTorch \cite{PyTorch} and PyTorch Geometric \cite{PyG} and run on a single Nvidia Quadro RTX 8000 GPU. For more details, we refer to our publicly available codebase \footnote{\url{https://github.com/lmu-dbs/diffusal}}.

\noindent\textbf{Competitors.~}
We compare DiffusAL with \emph{random} sampling, \emph{entropy} sampling~\cite{SettlesAL}, and \emph{coreset}~\cite{Core-Set} as graph-independent uncertainty-aware and diversity-aware active learning strategies, respectively. Furthermore, we include \textit{degree} sampling as a graph-based representativeness-based baseline, selecting the highest degree nodes, as well as the state-of-the-art graph-specific active learning methods \textit{AGE} \cite{AGE}, \textit{FeatProp}~\cite{FeatProp}, \textit{LSCALE}~\cite{LSCALE} and \textit{GRAIN} \cite{GRAIN}.

As proposed in~\cite{AGE,FeatProp,LSCALE,GRAIN}, all baselines use GCNs as classifiers, except LSCALE, which uses the proposed distance-based classifier. Our proposed method DiffusAL uses the introduced QBC as a classifier, and we provide comprehensive experiments showing the influence of the prediction model. 

\noindent\textbf{Hyperparameters.~}
We use the same hyper-parameters having a hidden layer size of 16, a dropout rate of 0.5, a learning rate of 0.01, and L2-regularization of $5 \times 10^{-4}$ as proposed in \cite{FeatProp}. For DiffusAL, we select $\alpha$ and $\epsilon$ as suggested in \cite{PushNet}.
We follow a batch selection and retrain from scratch after each acquisition round. However, to ensure more diverse uncertainties (and because the other two scores are static), we follow the setting of \cite{AGE} and also incrementally train the model for one epoch between instance selection within one acquisition round. The evaluation in \Cref{subsec:r3} shows that this does not impair our efficiency.
To provide a meaningful evaluation without the effects of an under-trained model or randomness factors, we report test accuracy for all approaches using a validation set of size $500$ and early stopping. However, the validation set is only part of the evaluation, not the procedure itself. We set the size of the initial pool to 2C (cf. \ref{subsec:preliminaries_notation}) and report results up to a budget of 20C with step sizes also twice the number of classes. To simulate a fairly realistic active learning scenario, the initial pool is sampled randomly without guaranteeing class balance for the baseline approaches without a specific initialization method. All experiments report an average of ten random seeds.

\begin{figure}[t]
	\begin{center}
    \begin{subfigure}{0.9\textwidth}
        \includegraphics[width=\textwidth]{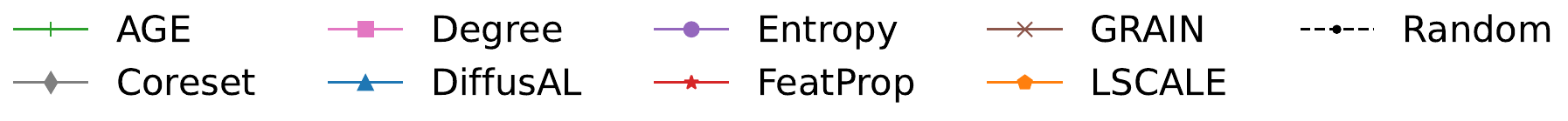}
    \end{subfigure}
        \begin{subfigure}{0.32\textwidth}
            \includegraphics[width=\textwidth]{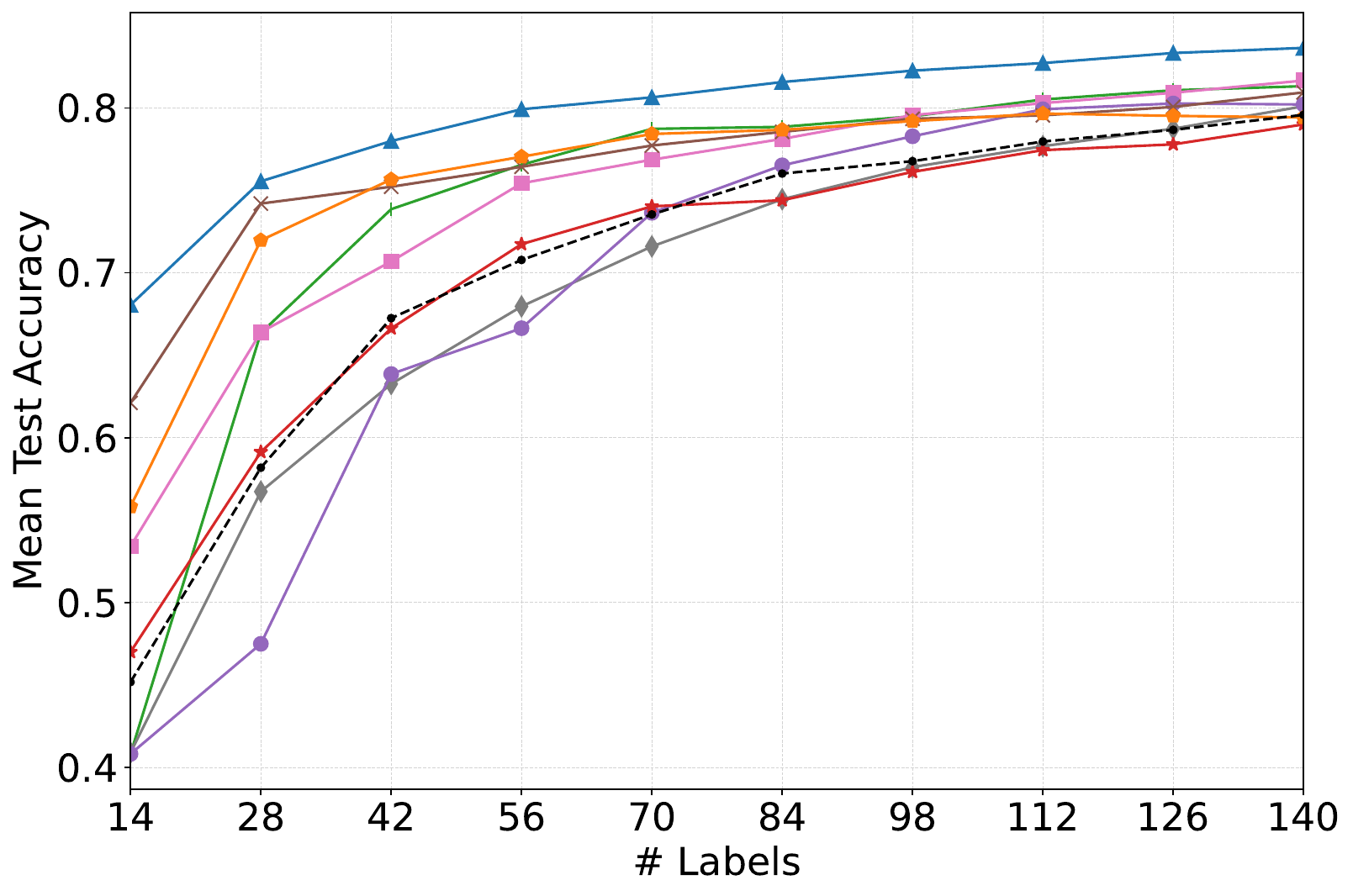}
            \caption{Cora.}
            \label{subfig:cora}
        \end{subfigure}
        \hfill        
        \begin{subfigure}{0.32\textwidth}
            \includegraphics[width=\textwidth]{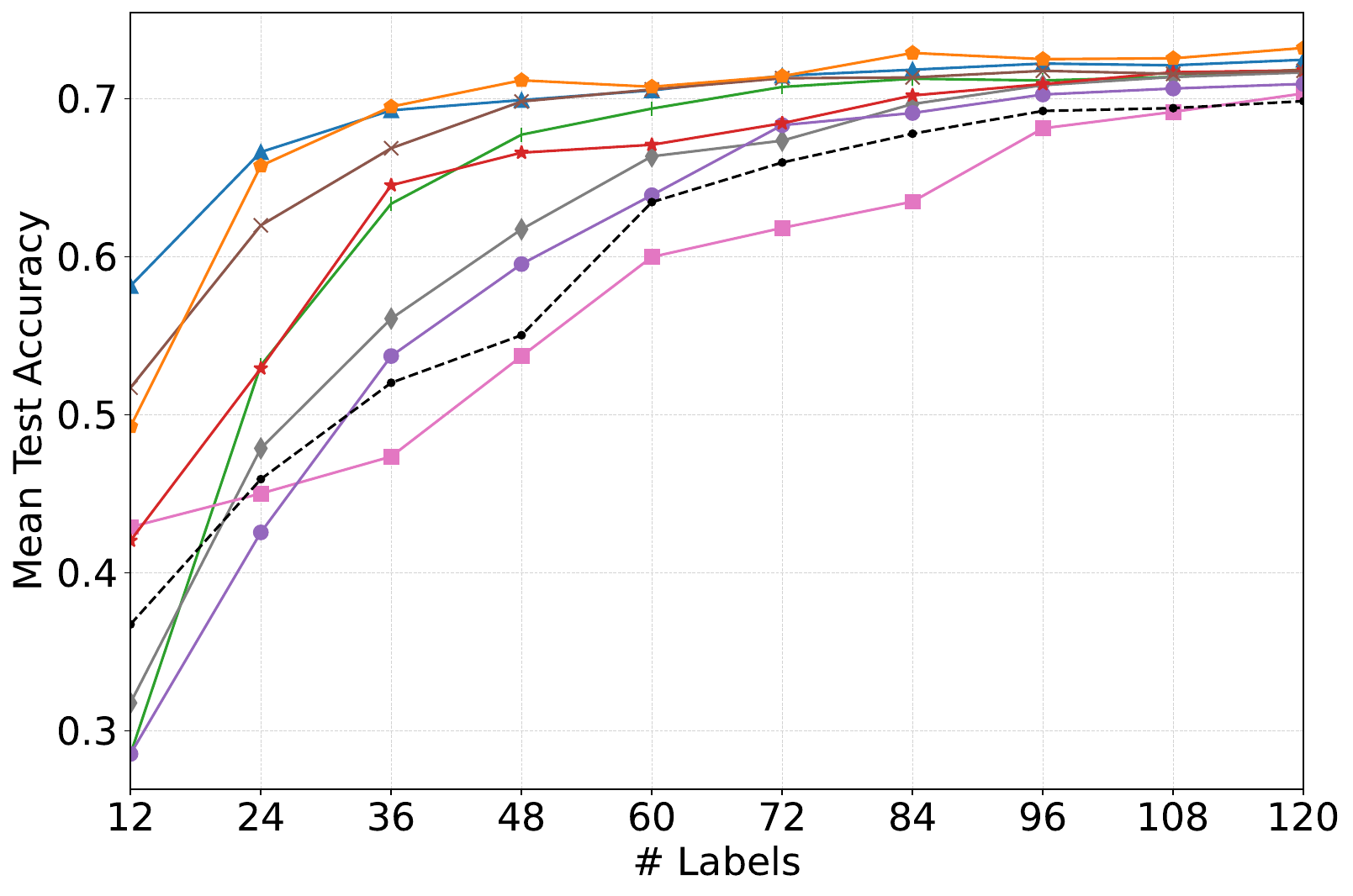}
            \caption{Citeseer.}
            \label{subfig:citeseer}
        \end{subfigure}
        \hfill
        \begin{subfigure}{0.32\textwidth}
            \includegraphics[width=\textwidth]{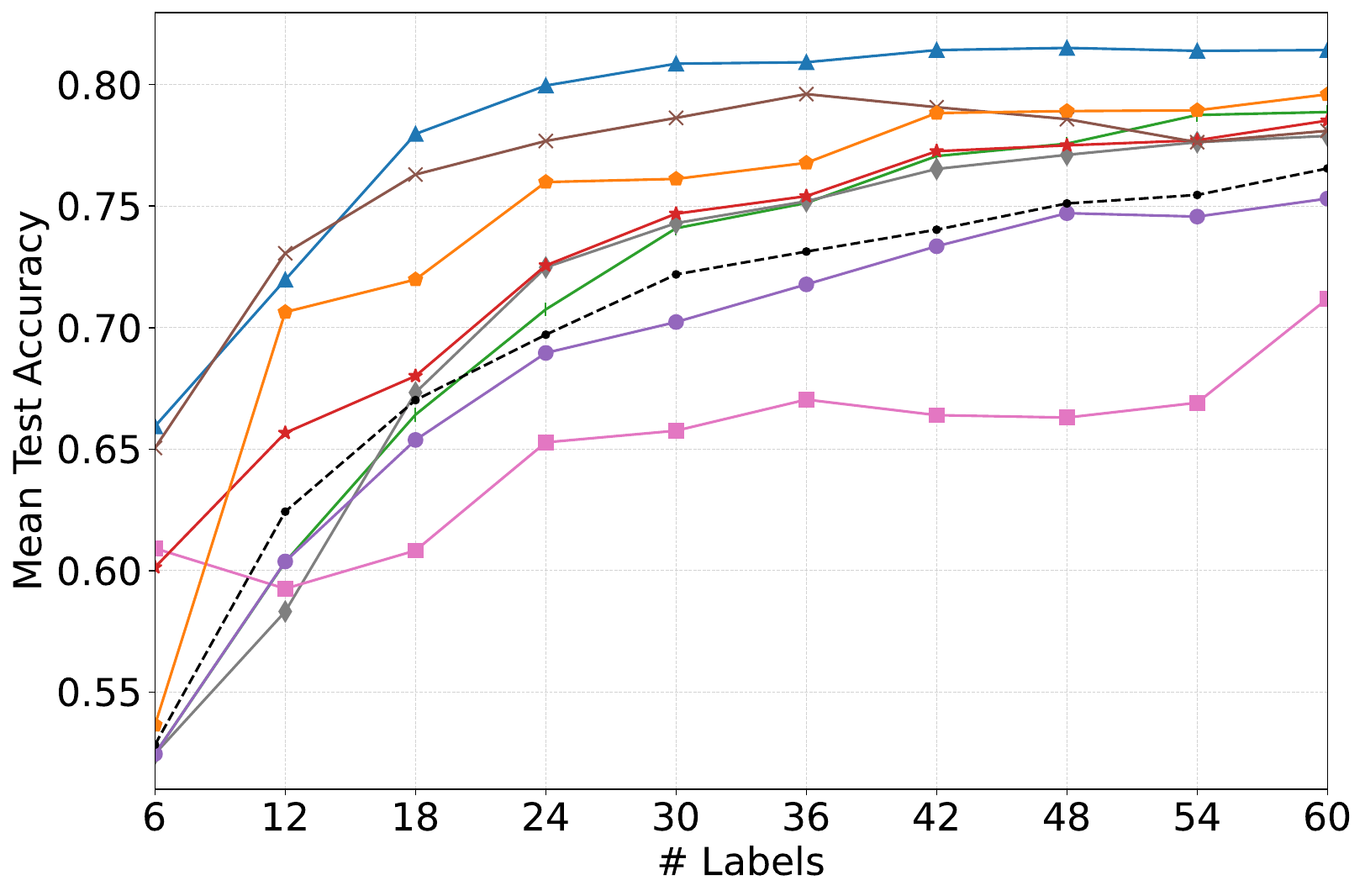}
            \caption{Pubmed.}
            \label{subfig:pubmed}
        \end{subfigure}
    \begin{subfigure}{0.32\textwidth}
      \includegraphics[width=\textwidth]{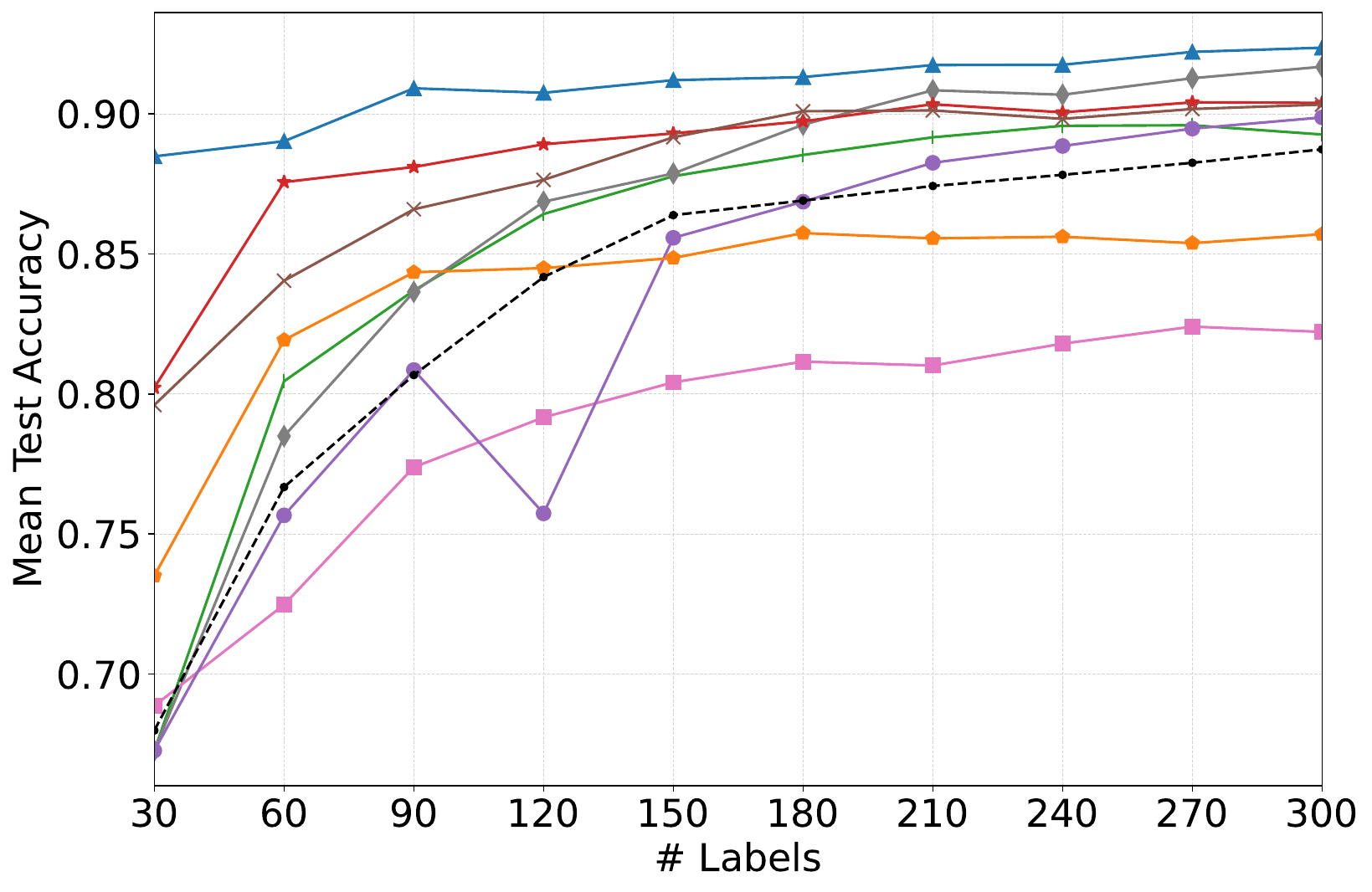}
        \caption{Coauthor-CS.}
        \label{subfig:cs}
    \end{subfigure}
    \hspace{1cm}
    \begin{subfigure}{0.32\textwidth}
        \includegraphics[width=\textwidth]{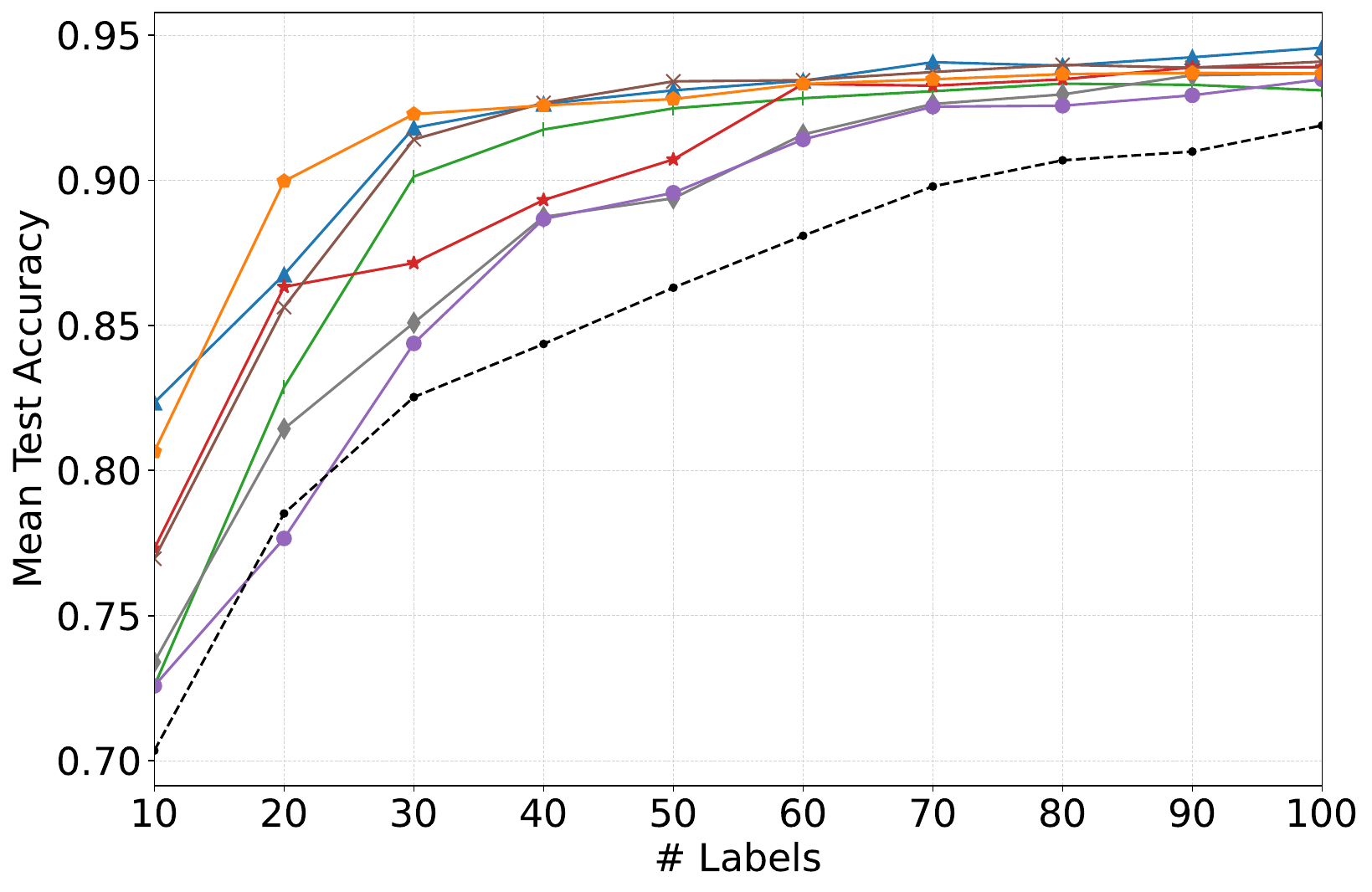}
        \caption{Coauthor-Physics.}
        \label{subfig:physics}
    \end{subfigure}
	\end{center}
	\caption{Active learning curves with the number of labeled nodes on the x-axis and average accuracy (over 10 random seeds) on the y-axis.}
	\label{fig:curves}
\end{figure}

\subsection{R1 - Performance Comparison}
\Cref{fig:curves} depicts the active learning curves for all budgets and datasets. 
DiffusAL (blue) is among the best-performing methods on all datasets. 
Especially on Cora and Coauthor-CS, we reach the highest mean accuracy for all labeling budgets and are the only competitor to reach a final accuracy of 83.6\% and 92.4\%, respectively. 
On Pubmed, GRAIN is similarly strong for the first two iterations. However, afterward, DiffusAL outperforms all methods for the remaining budgets and reaches a final average accuracy of 81.4\%. In comparison, LSCALE, the second-best performing method with respect to the final budget, only reaches 79.9\%.

\begin{wrapfigure}{r}{0.55\textwidth}
    \centering
    \includegraphics[width=0.55\textwidth]{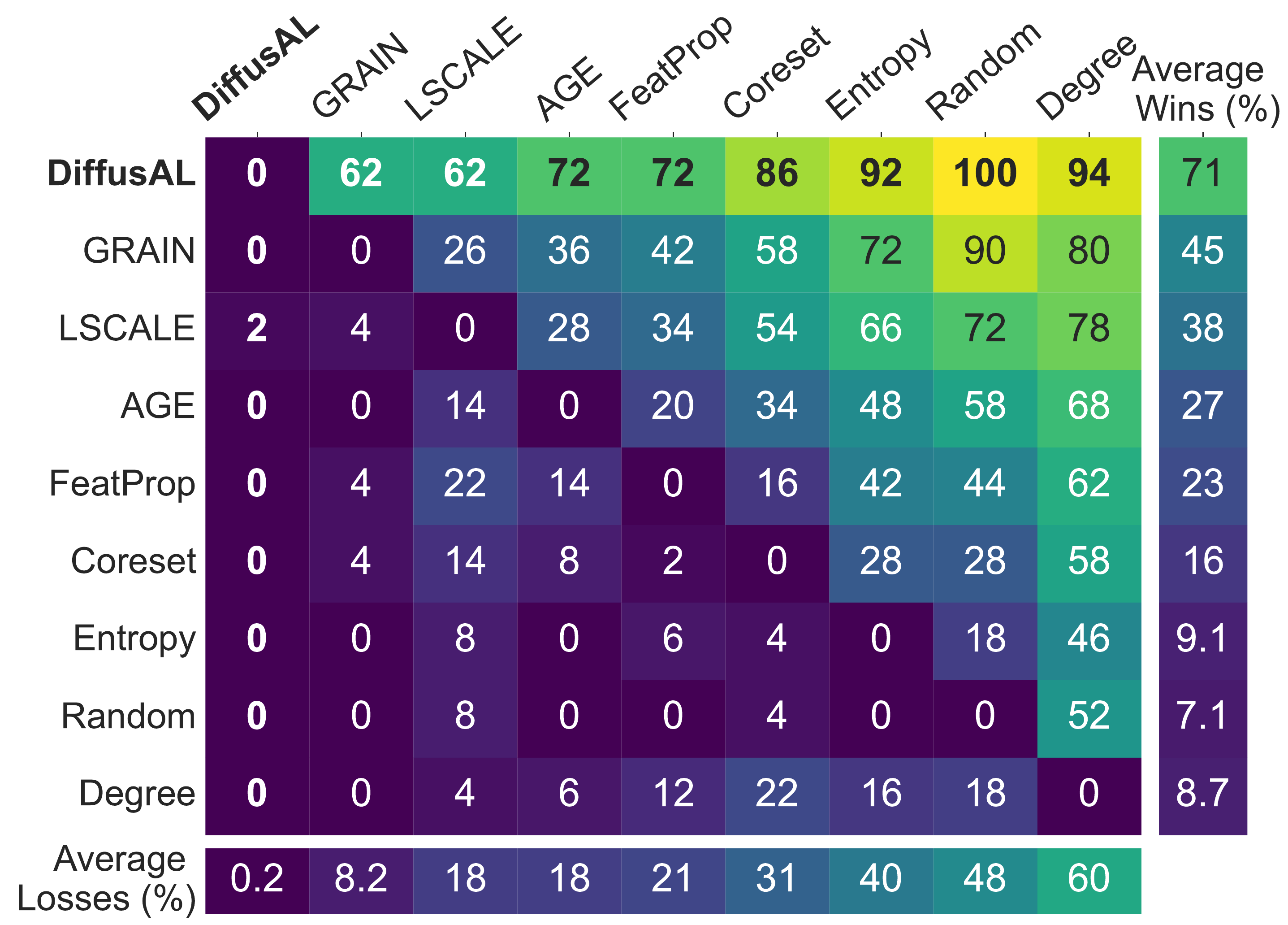}
    \caption{Pairwise dueling matrix. Cell $ij$ indicates how often competitor $i$ won against competitor $j$ \textbf{with statistical significance} over all datasets and labeling budgets (in \%). The bottom-most row and right-most column denote each method's average losses and wins, respectively (in \%). 
    }
    \label{fig:duelling} 
\end{wrapfigure}

On Citeseer and Physics\footnote{On Physics, Degree underperformed considerably and is therefore omitted for better presentation.}, GRAIN and LSCALE are similarly strong as DiffusAL. 
For both datasets, the learning curves converge to similar accuracies above a certain labeling budget for some methods such that a clear winner can no longer be pronounced. 
Therefore, \Cref{fig:duelling} provides a comprehensive dueling matrix indicating how often each strategy has won and lost against the other strategy in a similar fashion as was proposed in~\cite{BADGE}. 
We apply a two-sided t-test with a p-value of $0.05$ to the classification accuracies over 10 random seeds to count whether one method outperformed another with statistical significance.
In total, we evaluated 50 experimental settings for each strategy (5 different datasets, 10 different labeling budgets from 2C to 20C). The values in a column and row of a method denote the percentage of losses and wins against another method, respectively. The bottom row indicates the average losses of each strategy over all experiments, and the right-most column indicates the average wins of a strategy over all experiments. 
The losses and wins in the cells $c_{ij}$ and $c_{ji}$ do not necessarily add up to 100\%. The margin between the wins in cell $ij$ and the losses in cell $ji$ indicates how often the strategy $i$ has performed equally well as competitor $j$.
Both numbers, the average losses \emph{and} the average wins, are particularly interesting when evaluating the success of an active learning method. 

In summary, the dueling matrix reveals the following insights:
\begin{itemize}
    \item DiffusAL has the \textbf{fewest losses (0.2\%, see first column)} and the \textbf{most wins (71\%, see first row)}.
    \item DiffusAL \textbf{wins over random sampling most often (100\%)}.
    \item Concerning wins over random sampling, GRAIN is the second-best method (90\%). However, DiffusAL statistically never loses against GRAIN. 
    \item The only strategy that can outperform DiffusAL is LSCALE. However, we beat LSCALE in 62\% of experiments and lost only 2\% of experiments.
\end{itemize}

\subsection{R2 - Analysis of Contributing Factors}
The selected datasets vary widely in terms of the number of nodes, edges, features, classes, and class distribution, making it difficult to develop an approach that can perform well across the spectrum. In the following, we analyze which components contribute most to DiffusAL's success and why it is so strong over a broad range of datasets.
\Cref{table:ablations} shows the performance of DiffusAL (bottom row) and DiffusAL when switching off individual parts of the acquisition function, i.e., the diversity component (D), the uncertainty score (U) and the importance score (I) and exchanging the model architecture (middle rows) for 2C, 6C, and 12C labeling budgets on all datasets where C is the number of classes. Red, bold numbers indicate the smallest accuracy, indicating the largest influence of a switched-off component, and blue, bold numbers indicate the highest accuracy. We exchange the classifier with a single network variant (MLP) and with a GCN taking the raw features as input instead of diffused features (GCN). Furthermore, we report results when using an additive score instead of a multiplicative score.

\begin{table}[t]
\centering
\caption{Comparison of DiffusAL with ablated variants. \textcolor{blue}{\textbf{Blue, bold}} numbers indicate the \textcolor{blue}{\textbf{highest, i.e. best,}} accuracy. \textcolor{red}{\textbf{Red, bold}} numbers indicate the \textcolor{red}{\textbf{lowest, i.e. worst,}} accuracy and hence the component with largest influence.}
\resizebox{\textwidth}{!}{%
\begin{tabular}{ccc|ccc|ccc|ccc|ccc|ccc}
\toprule
 & & & \multicolumn{3}{c}{\textbf{Cora}} & \multicolumn{3}{c}{\textbf{Citeseer}} & \multicolumn{3}{c}{\textbf{Pubmed}} & \multicolumn{3}{c}{\textbf{CS}} & \multicolumn{3}{c}{\textbf{Physics}} \\
D & U & I &        2C &     6C &    12C &        2C &     6C &    12C &      2C &     6C &    12C &     2C &     6C &    12C &       2C &     6C &    12C \\

\midrule
&  &  \checkmark     &     \textcolor{red}{\textbf{45.5}} &  77.8 &  80.6 &     \textcolor{red}{\textbf{43.3}} &  \textcolor{red}{\textbf{63.8}} &  \textcolor{red}{\textbf{69.7}} &   \textcolor{red}{\textbf{56.3}} &  68.0 &  69.8 &  \textcolor{red}{\textbf{71.9}} &  \textcolor{red}{\textbf{81.7}} &  \textcolor{red}{\textbf{82.2}} &       \textcolor{red}{\textbf{71.9}} &  89.8 &  93.8 \\

 & \checkmark &  &     \textcolor{red}{\textbf{45.5}} &  76.1 &  80.1 &     \textcolor{red}{\textbf{43.3}} &  65.5 &  70.0 &   \textcolor{red}{\textbf{56.3}} &  \textcolor{red}{\textbf{70.6}} &  \textcolor{red}{\textbf{75.4}} &  \textcolor{red}{\textbf{71.9}} &  83.3 &  90.8 &    \textcolor{red}{\textbf{71.9}} &  89.6 &  93.1 \\   
 
 & \checkmark & \checkmark  &     \textcolor{red}{\textbf{45.5}} &  78.5 &  81.7 &     \textcolor{red}{\textbf{43.3}} &  69.8 &  71.3 &   \textcolor{red}{\textbf{56.3}} &  75.3 &  80.0 &  \textcolor{red}{\textbf{71.9}} &  89.3 &  91.4 &    \textcolor{red}{\textbf{71.9}} &  92.4 &  93.9 \\

 \checkmark & &       &     - &  \textcolor{red}{\textbf{74.5}} &  \textcolor{red}{\textbf{76.0}} &     - &  67.6 &  71.1 &   - &  64.6 &  76.5 &  - &  89.4 &  90.4 &       -&  \textcolor{red}{\textbf{86.4}} &  \textcolor{red}{\textbf{87.1}}  \\

\checkmark & & \checkmark &        - &  76.4 &  80.5 &        - &  67.7 &  71.0 &      - &  77.2 &  79.9 &     - &  87.5 &  87.3 &       - &  91.5 &  92.4 \\

\checkmark & \checkmark & &        - &  78.6 &  81.9 &    - &  69.1 &  71.0 &      - &  74.9 &  77.1 &     - &  90.5 &  91.6 &       - &  88.3 &  90.9 \\

\midrule

\multicolumn{3}{c}{\textbf{Additive}} &     - &  78.8 &  81.3 &       - &  \textcolor{blue}{\textbf{70.8}} &  71.3 &      - &  79.1 &  80.2 &  - &  \textcolor{blue}{\textbf{91.0}} &  \textcolor{blue}{\textbf{92.1}} &    - &  91.7 &  92.7 \\
\midrule
\multicolumn{3}{c}{\textbf{MLP}}            &     62.0 &  78.8 &  81.8 &     52.7 &  70.6 &  \textcolor{blue}{\textbf{71.8}} &   64.1 &  78.8 &  79.9 &  87.8 &  90.4 &  91.2 &    80.4 &  91.4 &  93.6 \\
\multicolumn{3}{c}{\textbf{GCN}}           &     61.8 &  77.5 &  80.7 &     49.8 &  69.3 &  71.3 &   64.5 &  76.5 &  78.5 &  83.3 &  89.6 &  91.2 &     \textcolor{blue}{\textbf{82.7}} &  91.6 &  93.1 \\

\midrule
\multicolumn{3}{c}{\textbf{DiffusAL}}       &     \textcolor{blue}{\textbf{68.0}} &  \textcolor{blue}{\textbf{79.9}} &  \textcolor{blue}{\textbf{82.3}} &     \textcolor{blue}{\textbf{58.2}} &  69.9 &  \textcolor{blue}{\textbf{71.8}} &   \textcolor{blue}{\textbf{65.9}} &  \textcolor{blue}{\textbf{80.0}} &  \textcolor{blue}{\textbf{81.4}} &  \textcolor{blue}{\textbf{88.5}} &  90.8 &  91.8 &    82.3 &  \textcolor{blue}{\textbf{92.6}} & \textcolor{blue}{\textbf{94.1}} \\ 

\bottomrule
\end{tabular}
}

\label{table:ablations}
\end{table}

The importance, uncertainty, and additive scoring have no influence on the initial pool selection, so we leave out numbers there. 
Our QBC robustifies the accuracy, especially in the first iteration, compared to the other two variants (MLP, GCN). The performance difference between the models gets smaller with increasing label information. In particular, when label information is sparse, the committee stabilizes the prediction. However, the diversity component has the largest impact on the initial set for all datasets. When switching off diversity (first three rows), the accuracy drops between 9.6\% (Pubmed) and 22.5\% (Cora). Other approaches, such as FeatProp or LSCALE, also use clustering in the first iteration. However, our sampling directly operates on the diffused features, which subsequently directly influence the training and thus results in a very strong initial performance.  

In general, switching off two scores yields worse results than only switching off one score, which indicates that the other two scores stabilize the results. But there is not one most important score over all datasets, supporting our claim that a robust selection benefits from diverse criteria. For instance, the accuracy drops the most when switching off \emph{uncertainty} and \emph{diversity} on Citeseer (by 2.1\%) and especially on Coauthor-CS (by 9.6\%). However, the performance on Cora and Physics primarily needs \emph{uncertainty} and \emph{importance}. In contrast, Pubmed benefits most from \emph{diversity} and \emph{importance}. 
Interestingly, some of our findings might give an indication of the performance of other methods. For instance, we found that importance, i.e., representativeness, is not beneficial on Coauthor-CS. LSCALE, which mainly focuses on representativeness sampling, yields the worst performance on this dataset. On Pubmed, however, uncertainty seems not to work well. Entropy and AGE both include uncertainty sampling and yield worse results. On Cora, where uncertainty and representativeness seem effective, Coreset and FeatProp, which mainly focus on diversity, are among the worst-performing methods.

Using an \emph{additive} score instead of a multiplicative score yields slightly worse results in general. From 10 comparisons, summing up the scores only yields three times slightly better results. However, the maximum difference is 0.9\% (Citeseer 6C), whereas using the multiplicative in DiffusAL, the additive score is up to 1.4\% (Physics 12C) better.

\begin{figure}[t]
    \centering
    \includegraphics[width=\textwidth,keepaspectratio]{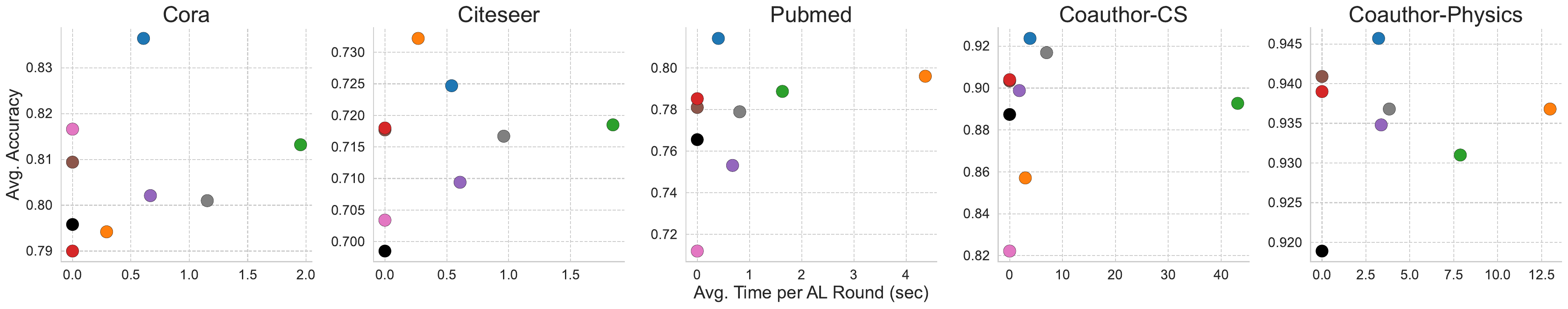}
    \includegraphics[width=\textwidth,keepaspectratio]{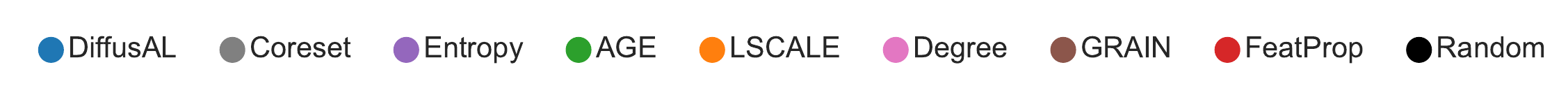}
    \caption[Time plot]{Average time in seconds (x-axis) required for one active learning round compared to the average final accuracy (y-axis) for all methods (color).}
    \label{fig:time_plot}
\end{figure}

\begin{table}[t]
\caption{Average time in seconds required for acquisition (acq), training (train), and in total ($\sum$) within one active learning iteration. Bold and underlined numbers indicate the fastest and second fastest methods, respectively. In total, DiffusAL is the fastest method on Physics and Pubmed, and the second fastest method on Cora and Citeseer.}
\centering
\resizebox{\textwidth}{!}{%
\begin{tabular}{lccccccccccccccc}
\toprule
 & \multicolumn{3}{c}{CS} & \multicolumn{3}{c}{Citeseer} & \multicolumn{3}{c}{Cora} & \multicolumn{3}{c}{Physics} & \multicolumn{3}{c}{Pubmed} \\
 &     acq & train & $\sum$ &      acq & train & $\sum$ &   acq & train & $\sum$ &     acq & train & $\sum$ &    acq & train & $\sum$ \\
\midrule
\textbf{AGE}             &  41.271 & 1.849 & 43.120 &    1.177 & 0.665 &  1.842 & 1.409 & 0.544 &  1.953 &   4.506 & 3.366 &  7.873 &  0.952 & 0.679 &  1.631 \\
\textbf{Coreset}         &   5.191 & 1.797 &  6.988 &    0.344 & 0.615 &  0.960 & 0.537 & 0.616 &  1.154 &   0.572 & 3.258 &  3.830 &  0.138 & 0.675 &  0.813 \\

\textbf{Entropy} &   \textbf{0.005} & 1.831 &  \textbf{1.836} &    \textbf{0.002} & 0.605 &  0.607 & \textbf{0.002} & 0.665 &  0.667 &   \textbf{0.011} & 3.358 &  \underline{3.369} &  \textbf{0.002} & 0.674 & \underline{ 0.676} \\
\textbf{LSCALE}          &   2.649 & \textbf{0.317} &  \underline{2.966} &    \underline{0.019} & \textbf{0.249} &  \textbf{0.269} & \textbf{0.042} & \textbf{0.250} &  \textbf{0.292} &  12.722 & \textbf{0.258} & 12.980 &  4.121 & \textbf{0.247} &  4.368 \\
\midrule
\textbf{DiffusAL}  &   \underline{2.567} & \underline{1.282} &  3.849 &    0.183 & \underline{0.356} &  \underline{0.539} & 0.268 & \underline{0.339} &  \underline{0.608} &   \underline{0.357} & \underline{2.863} &  \textbf{3.220} &  \underline{0.043} & \underline{0.361} &  \textbf{0.404} \\
\bottomrule
\end{tabular}
}
\label{tab:time_table}
\end{table}

\subsection{R3 - Acquisition and Training Efficiency}
\label{subsec:r3}
\Cref{fig:time_plot} shows the total average time (in seconds) for one active learning step on the x-axis (smaller is better) and the final accuracy after all 20C labels are selected on the y-axis (larger is better) for all methods (color). 

We focus on an iterative AL selection where re-training between acquisition steps is necessary to get new uncertainty scores. In contrast, GRAIN, FeatProp, degree sampling, and random sampling select all instances for labeling at once and do not require re-training. Therefore, their average time is set to zero, and their accuracy is plotted for comparison. 
However, these methods are generally less label-efficient since they are not directly coupled to the current learning model. Except for Citeseer, DiffusAL is always on the Pareto-front, yielding the best final average accuracy while still being fairly time-efficient.
In \Cref{tab:time_table}, we split the total time into the acquisition and the training time for the iterative methods. All GCN-based methods (Coreset, AGE, Entropy) denote fairly similar training times. Despite using an ensemble, DiffusAL is slightly faster than the GCN-based methods since the features are pre-computed. AGE and Coreset both require a longer time for acquisition. AGE can exploit pre-calculated centrality scores. However, the uncertainty score and especially the density score must be freshly calculated in each round. Especially for the very large graph data CS, AGE requires over 40 seconds for one active learning iteration. 
Coreset extracts the latent representations from the current model and requires the computation of a pairwise distance matrix. Compared to that, DiffusAL only needs to calculate the uncertainty scores derived from the QBC model since the other scores are pre-computed. 
Only the entropy-based selection scheme has a faster acquisition time since it only needs one forward pass through the network.

LSCALE, which also defined a dedicated network towards a unified learning and selection framework, has the fastest training times out of all methods. However, depending on the dataset, the acquisition time is much larger than DiffusAL's acquisition time. As such, the overall time needed for one active learning round varies considerably between datasets. For instance, on Citeseer and Cora, LSCALE is the fastest method out of all iterative methods. Still, on the much larger graphs Pubmed and Physics, it is the slowest method due to larger acquisition times (4.4 seconds and 12.7 seconds, respectively).
Overall, even though we use an ensemble method, our training and acquisition times are fairly stable across datasets and, in total, comparably good as plain uncertainty sampling with a GCN.

\section{Conclusion}
The annotation of unlabeled nodes in graphs is a time-consuming and costly task and, accordingly, it is of great interest to advance label-efficient methods. 
Motivated by the success of diffusion-based graph learning approaches, we propose DiffusAL, a novel active learning strategy for node classification. DiffusAL uses diffusion to predict node labels accurately and compute meaningful utility scores consisting of \emph{model uncertainty}, \emph{diffused feature diversity}, and \emph{node importance} for active node selection, such that training and data selection cooperate toward label-efficient node classification. 
DiffusAL is significantly better generalizable over a wide range of datasets and is, in terms of statistical significance, not beaten by any other method in 99.8\% of all experiments. Moreover, it is the only method that significantly outperforms random selection in 100\% of the evaluated settings. 
Due to pre-computed features stored in a diffusion matrix, our model can efficiently compute a node's utility for training and acquisition. 
Our extensive ablation study shows that each component of DiffusAL contributes to different datasets and active learning stages, making it robust in diverse graph settings.

%
%
\bibliographystyle{splncs04}
\bibliography{references}

%
%
%
%
%
%
%


\appendix 
\section{Diffusion and Node Importance}
We provide further analysis on the node importance scores since they are a key property of DiffusAL. 

\begin{figure}
    \begin{center}
        \begin{subfigure}{0.32\textwidth}
            \centering
            \includegraphics[width=\textwidth]{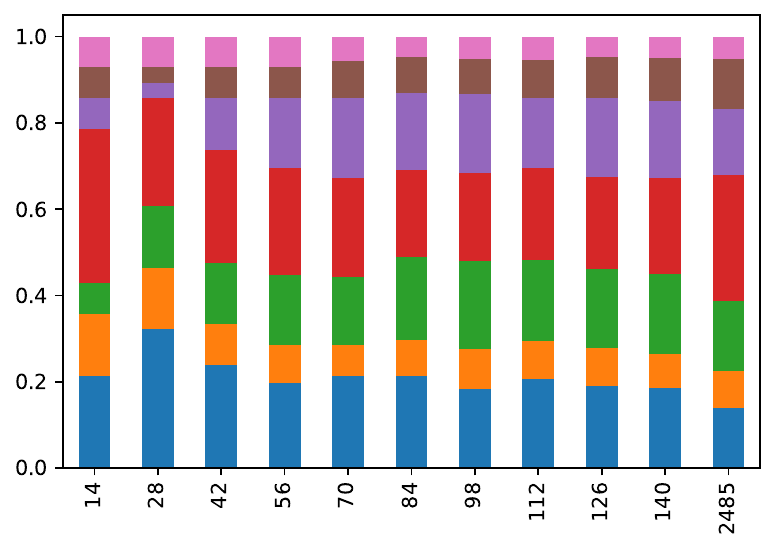}
            \caption{Cora.}
        \end{subfigure}
        \hfill
        \begin{subfigure}{0.32\textwidth}
            \centering
            \includegraphics[width=\textwidth]{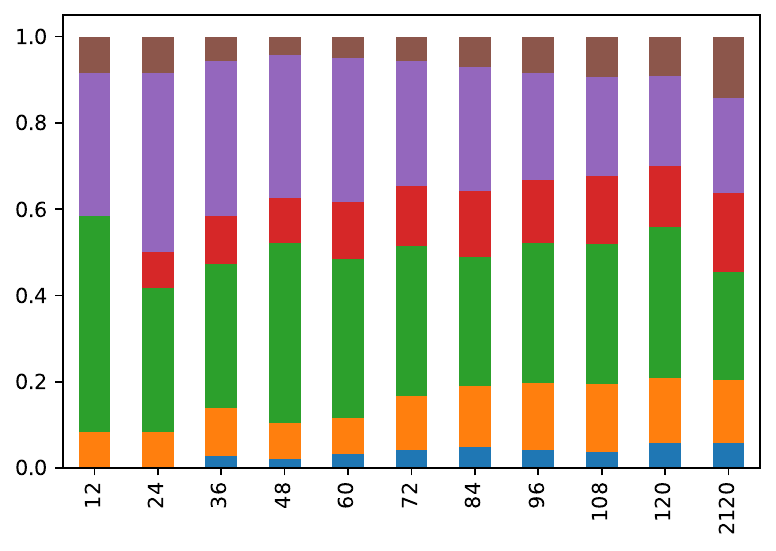}
            \caption{Citeseer.}
        \end{subfigure}
        \hfill
        \begin{subfigure}{0.32\textwidth}
            \centering
            \includegraphics[width=\textwidth]{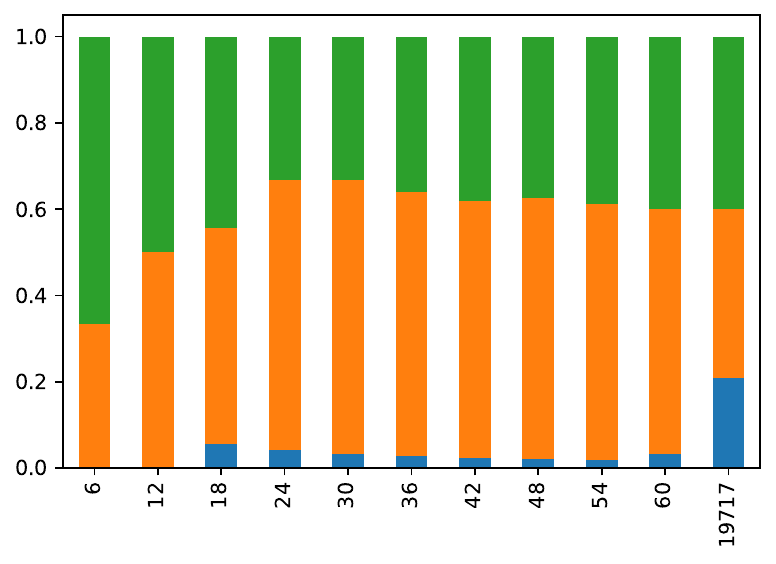}
            \caption{Pubmed.}
        \end{subfigure}

    \end{center}
    \caption{Class distribution of the most important nodes: the x-axis represents various budgets of nodes, the y-axis measures the fraction that each class makes up in the given labeled set. Colors indicate the respective classes.}
    \label{fig:class_dist_imp}
\end{figure}

\paragraph{Class Distribution of Important Nodes}
\Cref{fig:class_dist_imp} displays the class distribution of the top k most important nodes for the citation networks. The last bar indicates the original class distribution comprising all nodes. On Pubmed, the minority class is heavily underrepresented in the top 60 most important nodes, leading the node importance score in our active selection to favor samples from the other two majority classes. On Citeseer, a similar but weaker trend can be observed. In contrast, on Cora, the distribution of the top k most important nodes rapidly approximates the actual class distribution. 

\begin{figure}[t]
    \begin{center}
        \begin{subfigure}{0.32\textwidth}
            \centering
            \includegraphics[width=\textwidth]{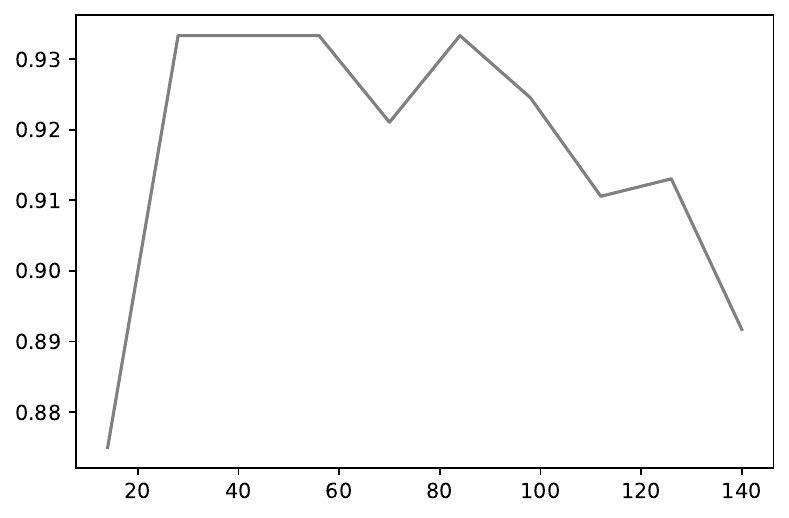}
            \caption{Cora.}
        \end{subfigure}
        \hfill
        \begin{subfigure}{0.32\textwidth}
            \centering
            \includegraphics[width=\textwidth]{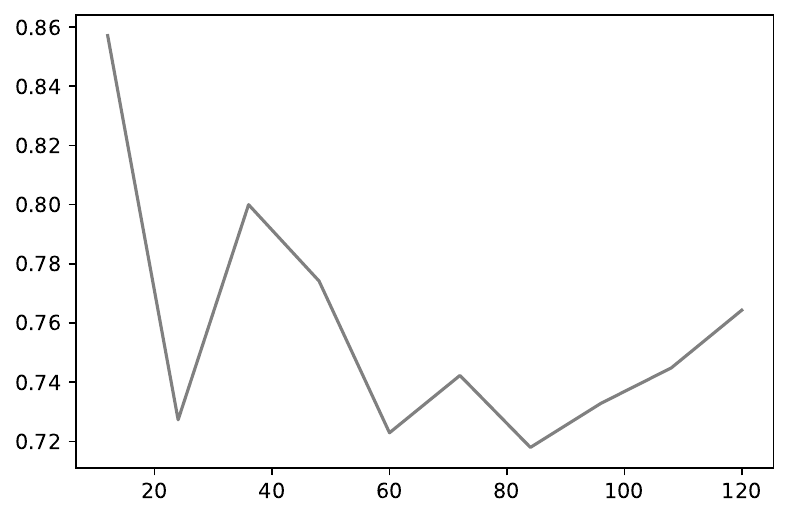}
            \caption{Citeseer.}
        \end{subfigure}
        \hfill
        \begin{subfigure}{0.32\textwidth}
            \centering
            \includegraphics[width=\textwidth]{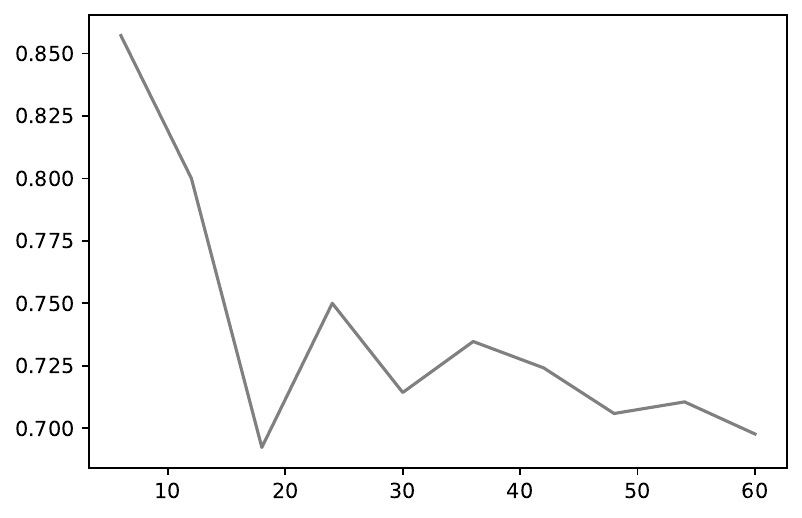}
            \caption{Pubmed.}
        \end{subfigure}

    \end{center}
    \caption{Overlap of most important and highest degree nodes for a given budget - x: various sampling budgets, y: the fraction of nodes appearing in both respective sets of nodes. }
    \label{fig:relationship_imp_degree}
\end{figure}

\paragraph{Node Importance vs. Degree}
As already discussed, the intuitive interpretation of the importance score for a given node \textit{i} is the probability of a random walk that starts at a random node \textit{j} to end at node \textit{i}. Consequently, a valid assumption is that important nodes tend to have an above-average degree since high-degree nodes are more likely to be visited during a random walk. Here, we analyze how the degree differs from node importance since common centrality measures, such as degree centrality, have already been established as active learning criteria in related work. \Cref{fig:relationship_imp_degree} displays the overlap of the most important nodes compared to the highest degree nodes for different budgets up to $20C$ for each dataset. Cora has the highest average overlap for the considered budgets with over $90\%$. However, for the other datasets, this overlap is a lot smaller at only around $75\%$ on average, indicating that node degree, while still seemingly a large one, is not the only influencing factor determining the importance of a node. Furthermore, a general insight is that the overlap for the topmost important nodes is the largest and decreases afterward. 

\paragraph{Node Importance and Diffusion vs. 2-hop}

\begin{figure}[t]
	\begin{center}
    \begin{subfigure}{0.32\textwidth}
        \includegraphics[width=\textwidth]{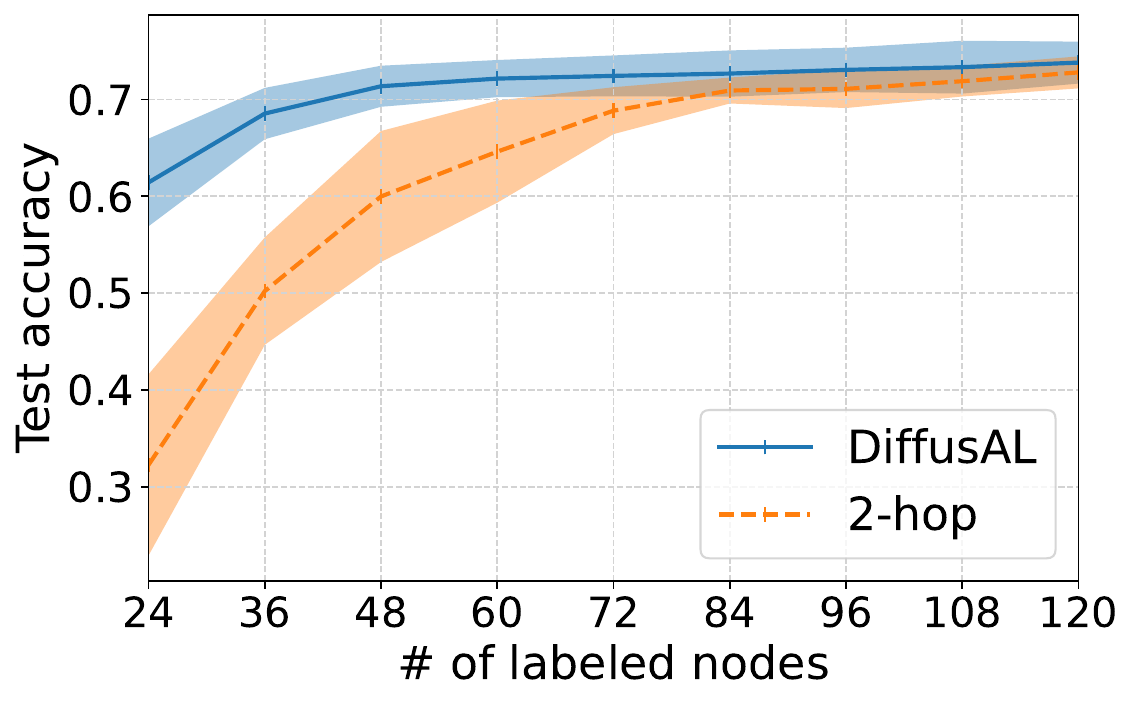}
        \caption{Citeseer.}
    \end{subfigure}
    \begin{subfigure}{0.32\textwidth}
        \includegraphics[width=\textwidth]{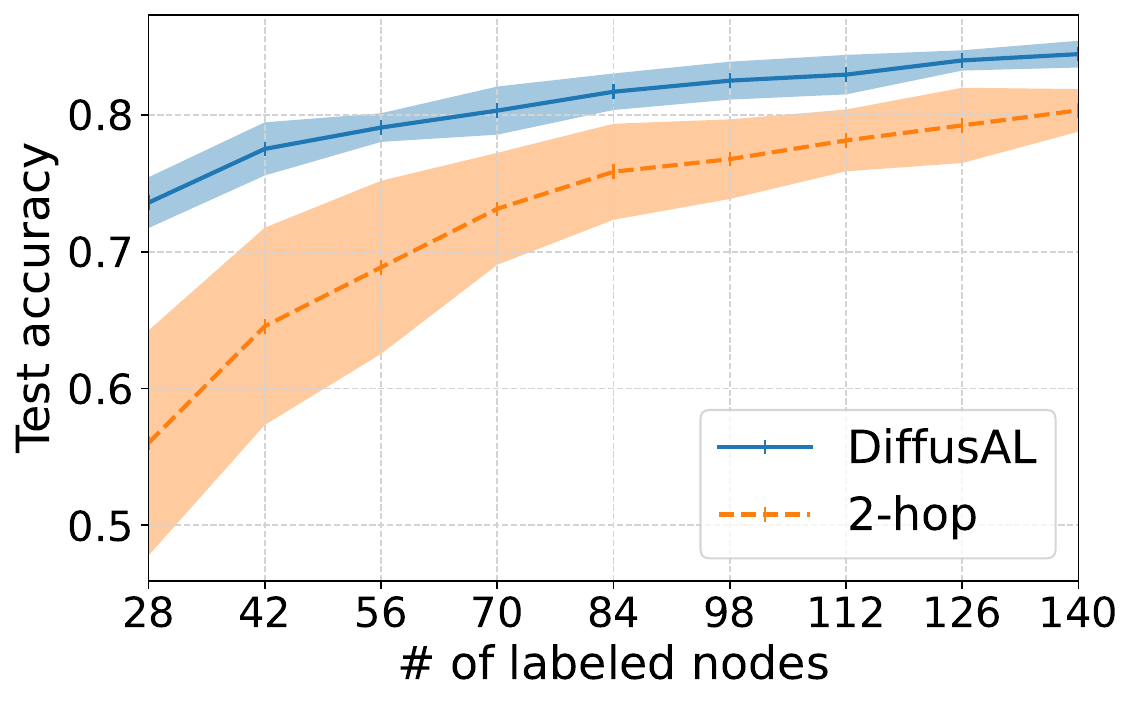}
        \caption{Cora.}
    \end{subfigure}
    \begin{subfigure}{0.32\textwidth}
        \includegraphics[width=\textwidth]{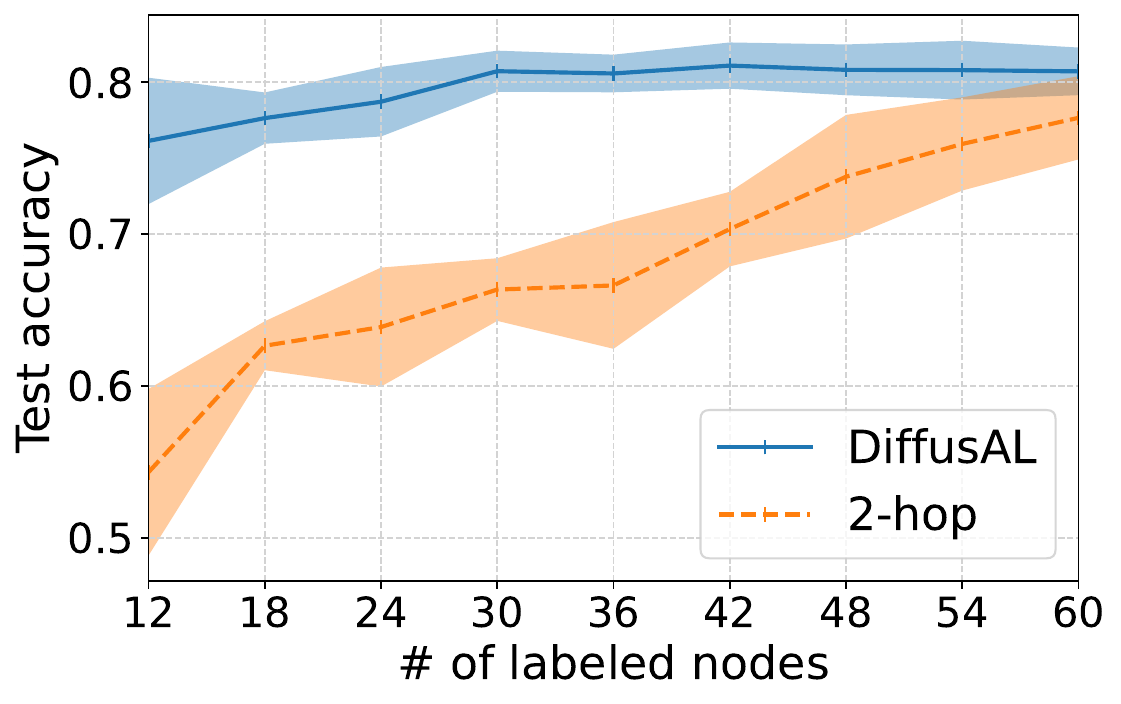}
        \caption{Pubmed.}
    \end{subfigure}
	\end{center}
	\caption{DiffusAL (blue) compared to a purely \textit{2}-hop-based alternative (orange).}
	\label{fig:ppr_2hop}
\end{figure}

We further analyze the influence of diffusion on sampling \emph{and} graph learning and drop all diffusion-related content to demonstrate the advantages. We replace every aspect concerned with diffusion with comparable components entirely based on \textit{k}-hop (2-hop) neighborhoods. Instead of using the PPR matrix to compute propagated features, the original adjacency matrix (squared and symmetrically normalized) is used. Furthermore, these features serve as the basis for the labeled pool initialization, cluster affiliation, and classifier training. Additionally, instead of the PPR matrix, we consider the column-wise sum of the adjacency matrix for the importance score. The results, depicted in \Cref{fig:ppr_2hop}, reveal the advantages of diffusion. Test accuracy decreases for all three citation networks when using the 2-hop-based replacement scores while variability increases, as can be seen from the wider error bands.

\section{Further Explanation on DiffusAL vs. AGE}
Among all competitors, AGE is arguably the most related one. However, there are some important differences regarding the concrete realization of the acquisition function. 
AGE uses a density score as well as a centrality score. Both are common choices for representativeness. As such, there might be a strong tendency to favor very representative instances from the graph. In contrast, we use a diversity score and node importance as representativeness estimate.  
The only purpose of the diversity score is to ensure that no region is oversampled. However, the node importance ensures that influential points are selected. 
A key difference to the centrality score used in AGE is that node importance not only considers the local neighborhood but takes the whole graph structure into account.
Furthermore, our selection and training both exploit expressive features in a consistent fashion. That is, the training directly makes use of the precomputed features. These features are also used for clustering, directly ensuring the diversity that is known to the model. The node importance directly corresponds to nodes that are most influential.  
Lastly, DiffusAL does not use any time-sensitive weighting parameters.

\section{Hyperparameter Variations}
We further have conducted experiments to show the robustness of DiffusAL regarding the hyper-parameters for model training.
For all datasets, DiffusAL denotes quite stable learning curves. Only on CS, hidden size, dropout, and weight decay seem to have a larger impact. 

\begin{figure}[t]
    \begin{center}
        \begin{subfigure}{\textwidth}
            \centering
            \includegraphics[width=\textwidth]{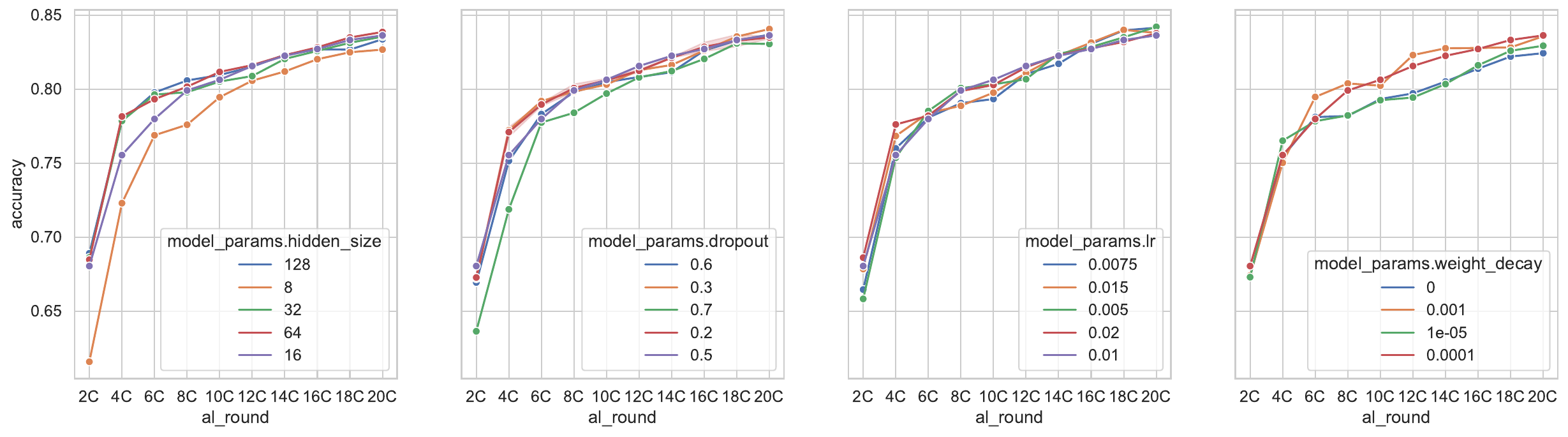}
            \caption{Cora.}
        \end{subfigure}
        \begin{subfigure}{\textwidth}
            \centering
            \includegraphics[width=\textwidth]{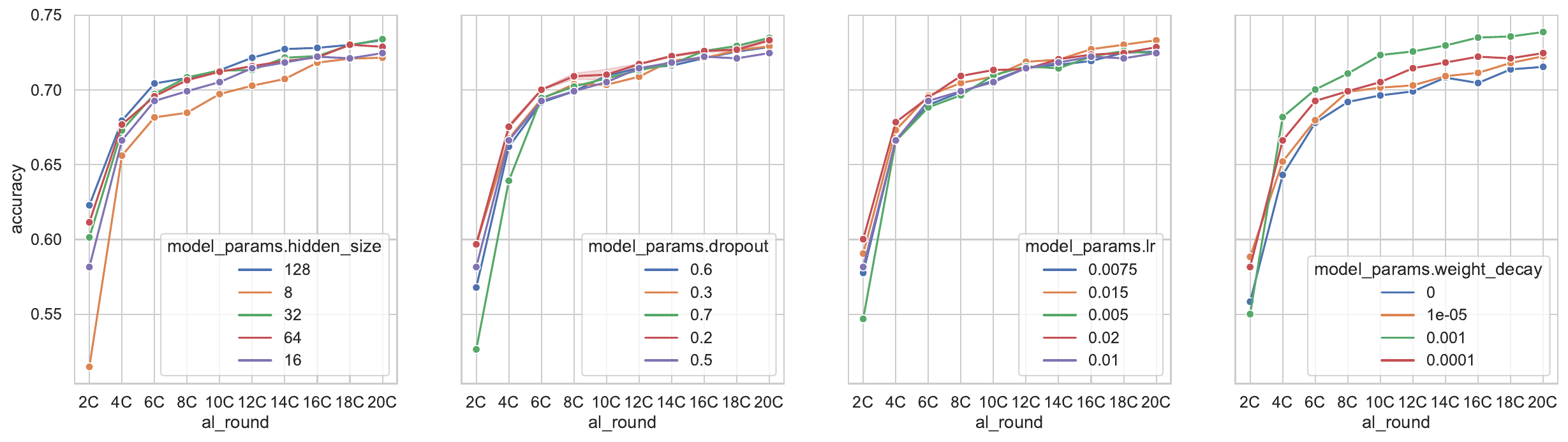}
            \caption{Citeseer.}
        \end{subfigure}
        \begin{subfigure}{\textwidth}
            \centering
            \includegraphics[width=\textwidth]{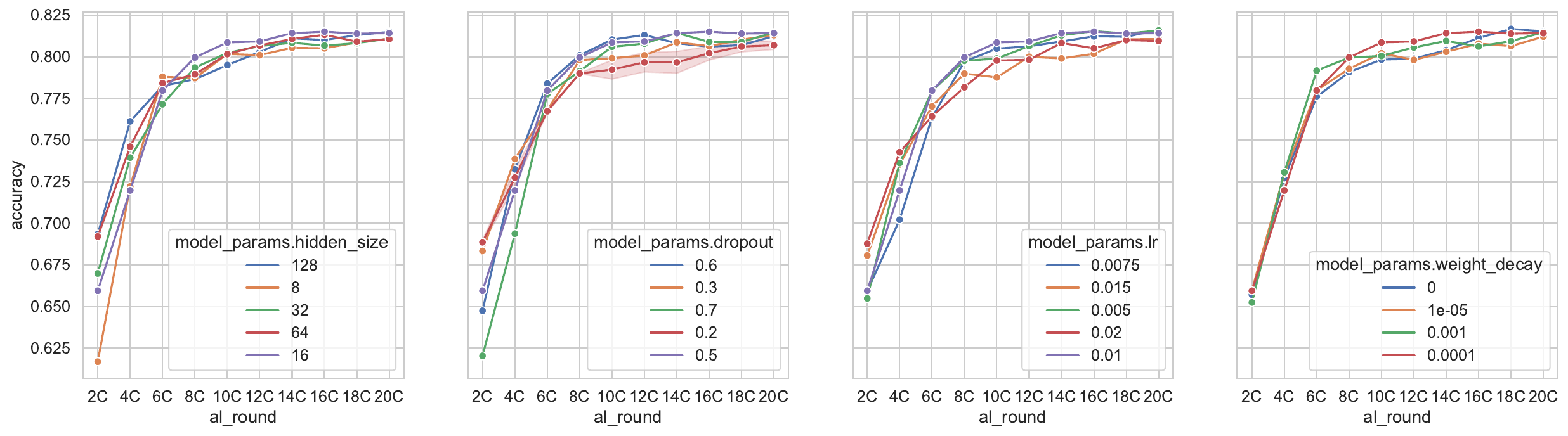}
            \caption{Pubmed.}
        \end{subfigure}
        \begin{subfigure}{\textwidth}
            \centering
            \includegraphics[width=\textwidth]{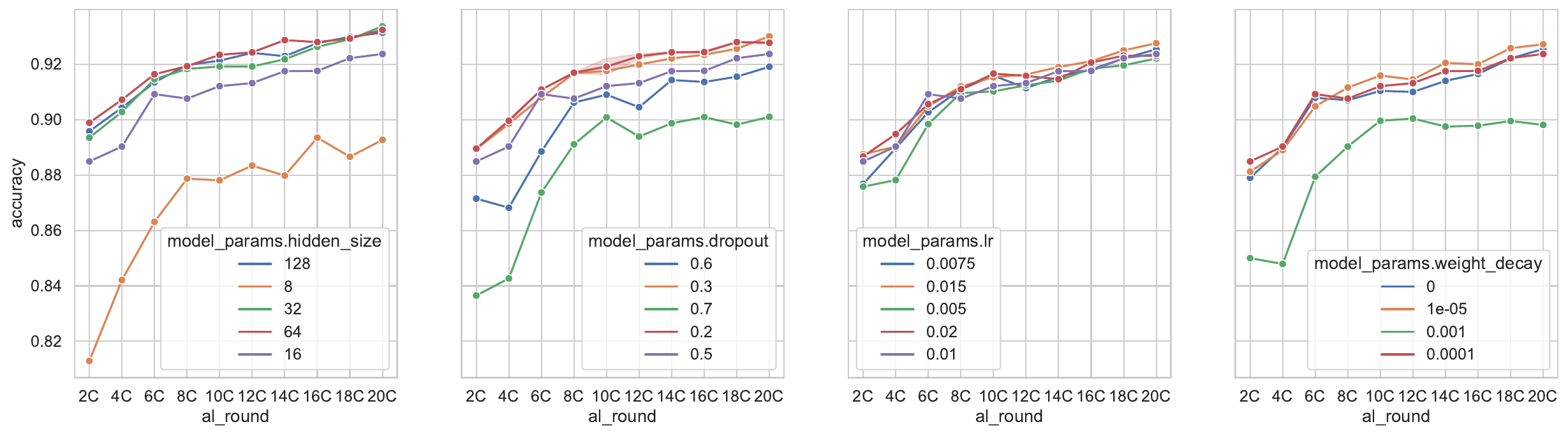}
            \caption{CS.}
        \end{subfigure}
        \begin{subfigure}{\textwidth}
            \centering
            \includegraphics[width=\textwidth]{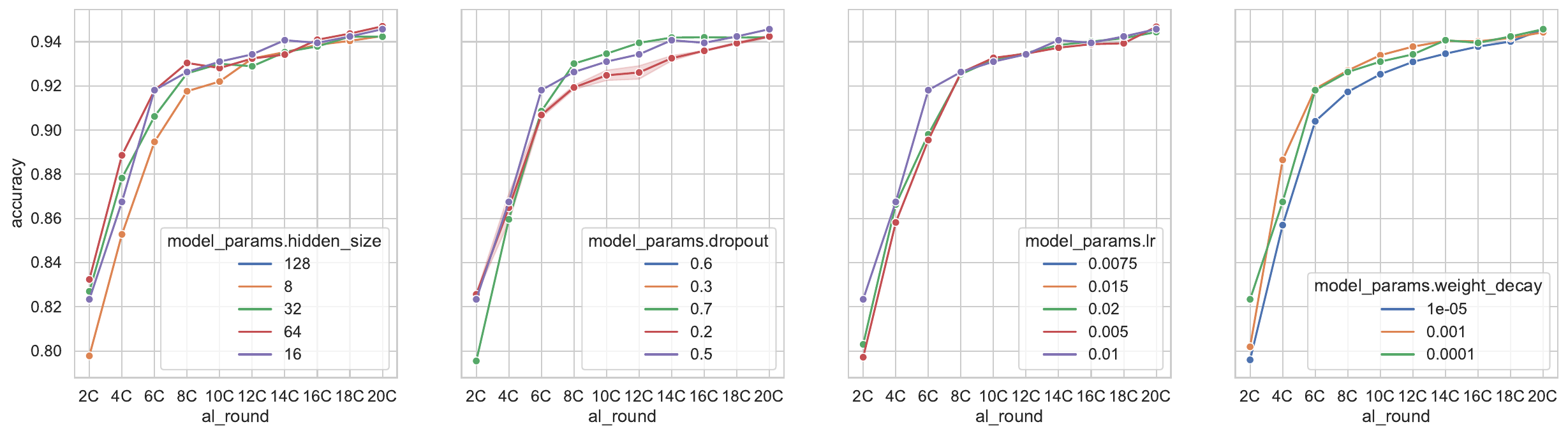}
            \caption{Physics.}
        \end{subfigure}

    \end{center}
    \caption{Hyperparameter variation. We show results when increasing and decreasing values for hidden size (column 1), dropout (column 2), learning rate (column 3), and weight decay (column 4) .}
    \label{fig:batch_sizes}
\end{figure}

\end{document}